\documentclass{bmvc2k}

\usepackage{amsmath,amssymb,xspace,floatrow}
\newcommand{\tss}[1]{\textsuperscript{#1}}
\newcommand{\app}{\raise.17ex\hbox{$\scriptstyle\sim$}}

\newcommand{\chk}{{\centering\checkmark}}
\newcommand{\Caption}[1]{\caption{\small#1}}
\newcommand{\AP}[1]{AP$^{#1}$\xspace}
\newlength\savewidth\newcommand\shline{\noalign{\global\savewidth\arrayrulewidth
 \global\arrayrulewidth 1pt}\hline\noalign{\global\arrayrulewidth\savewidth}}

\newcommand{\fig}[1]{Figure~\ref{fig:#1}}
\newcommand{\tab}[1]{Table~\ref{table:#1}}

\title{A MultiPath Network for Object Detection}
\addauthor{Sergey Zagoruyko$^\star$,~~Adam Lerer,~~Tsung-Yi Lin$^\star$,~~Pedro O.~Pinheiro$^\star$,\\\hbox{}\hspace{12mm}Sam Gross,~~Soumith Chintala,~~Piotr Doll\'ar\\~\\Facebook AI Research (FAIR)}{}{1}
\addinstitution{}
\runninghead{Zagoruyko \MakeLowercase{et al.}}{A MultiPath Network for Object Detection}
\begin{document}
\maketitle

\begin{abstract}
The recent COCO object detection dataset presents several new challenges for object detection. In particular, it contains objects at a broad range of scales, less prototypical images, and requires more precise localization. To address these challenges, we test three modifications to the standard Fast R-CNN object detector: (1) skip connections that give the detector access to features at multiple network layers, (2) a foveal structure to exploit object context at multiple object resolutions, and (3) an integral loss function and corresponding network adjustment that improve localization. The result of these modifications is that information can flow along multiple paths in our network, including through features from multiple network layers and from multiple object views. We refer to our modified classifier as a `MultiPath' network. We couple our MultiPath network with DeepMask object proposals, which are well suited for localization and small objects, and adapt our pipeline to predict segmentation masks in addition to bounding boxes. The combined system improves results over the baseline Fast R-CNN detector with Selective Search by 66\% overall and by $4\times$ on small objects. It placed second in both the COCO 2015 detection and segmentation challenges.
\end{abstract}

\section{Introduction}

Object classification~\cite{AlexNet, Simonyan15, GoogLeNet} and object detection~\cite{SermanetICLR2013, Szegedy15, girshick15fastrcnn} have rapidly progressed with advancements in convolutional neural networks (CNNs)~\cite{lecun1998gradient} and the advent of large visual recognition datasets~\cite{Everingham10, imagenet_cvpr09, mscoco2015}. Modern object detectors predominantly follow the paradigm established by Girshick et al. in their seminal work on Region CNNs~\cite{Girshick2014rcnn}: first an object proposal algorithm~\cite{Hosang2015proposals} generates candidate regions that may contain objects, second, a CNN classifies each proposal region. Most recent detectors follow this paradigm~\cite{gidaris2015object, girshick15fastrcnn, RenNIPS15fasterRCNN} and they have achieved rapid and impressive improvements in detection performance.

Except for concurrent work (e.g.~\cite{bell15ion, he2015deep, he2015segmentation}), most previous object detection work has focused on the PASCAL~\cite{Everingham10} and ImageNet~\cite{imagenet_cvpr09} detection datasets. Recently, the COCO dataset~\cite{mscoco2015} was introduced to push object detection to more challenging settings. The dataset contains 300,000 images of fully segmented object instance in 80 categories, with an average of 7 object instances per image. COCO introduces a number of new challenges compared to previous object detection datasets: (1) it contains objects at a broad range of scales, including a high percentage of small objects, (2) objects are less iconic, often in non-standard configurations and amid clutter or heavy occlusion, and (3) the evaluation metric encourages more accurate object localization.

\begin{figure}[t]
 \includegraphics[width=1\textwidth]{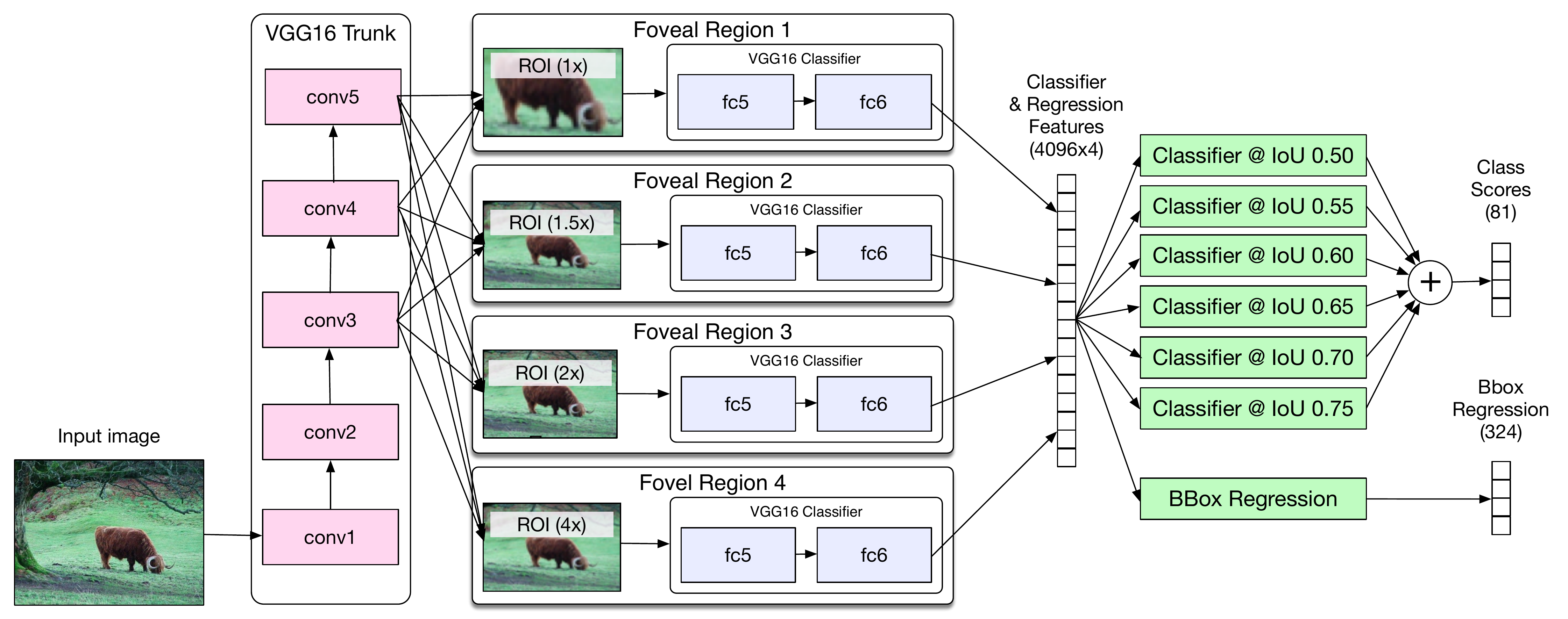}
\Caption{Proposed MultiPath architecture. The COCO dataset~\cite{mscoco2015} contains objects at multiple scales, in context and among clutter, and under frequent occlusion. Moreover, the COCO evaluation metric rewards high quality localization. To addresses these challenges, we propose the MultiPath network pictured above, which contains three key modifications: skip connections, foveal regions, and and an integral loss function. Together these modifications allow information to flow along multiple paths through the network, enabling the classifier to operate at multiple scales, utilize context effectively, and perform more precise object localization. Our MultiPath network, coupled with DeepMask object proposals~\cite{pinheiro2015learning, pinheiro2016refining}, achieves major gains on COCO detection.}
\label{fig:model}
\end{figure}

In this paper, we revisit recent improvements in object detection by performing extensive experiments on the COCO dataset. In particular, we begin with the Fast R-CNN object detector~\cite{girshick15fastrcnn}, and test a number of intuitive modifications to explicitly address the unique challenges of this dataset, including small object detection, detection of objects in context, and improved localization. Our goal is to adapt the highly successful Fast R-CNN object detector to perform better in these settings, and we use COCO to drive our experiments.

Inspired by recent advances in object detection, we implement three network modifications: (1) a multi-stage feature aggregator that implements skip connections in intermediate network layers to more accurately detect objects at multiple scales, (2) a foveal structure in the classifier network that helps improve localization by looking at multiple image contexts, and (3) a novel loss function and corresponding network adjustment that optimize an integral of localization overlaps and encourage higher-precision localization. These three modifications allow information to flow along multiple paths in our network, including through features from multiple network layers and from multiple object views, see \fig{model}. We therefore refer to our approach as a `MultiPath' network.

We train our MultiPath detector using the recently proposed DeepMask object proposals~\cite{pinheiro2015learning, pinheiro2016refining}, which, like our model, are well adapted to the COCO dataset. Our combined system, using DeepMask proposals and our MultiPath classifier, achieves a detection score of 33.5 average precision (AP) for detection with an ensemble of 6 models. Compared to the baseline Fast R-CNN detector~\cite{girshick15fastrcnn} with Selective Search proposals~\cite{Uijlings13}, which achieves an AP of 19.3, this represents a 66\% improvement in performance. Moreover, for small objects we improve AP by nearly 4$\times$. We also adopt our system to generate segmentation masks, and achieve an AP of 25.1 on the segmentation task.

Our system placed second in the 2015 COCO Detection Challenge in both the bounding box and segmentation tracks. Only the deeper ResNet classifier~\cite{he2015deep} outperformed our approach. Potentially, ResNet could be used as the feature extractor in our MultiPath network.

\section{Related Work}

Object detection is a fundamental and heavily-researched task in computer vision. Until recently, the sliding window paradigm was dominant~\cite{Viola2004, DollarPAMI14pyramids}, especially for face and pedestrian detection. Deformable part models~\cite{Felzenszwalb2010} followed this framework but allowed for more object variability and thus found success across general object categories; likewise, Sermanet et al.~\cite{SermanetICLR2013, sermanetCVPR2013} showcased the use of CNNs for general object detection in a sliding window fashion. More recent detectors follow the region-proposal paradigm established by Girshick et al.~\cite{Girshick2014rcnn} in which a CNN is used to classify regions generated by an object proposal algorithm~\cite{Hosang2015proposals}. Many recent detectors follow this setup~\cite{gidaris2015object, Szegedy15, He2014sppNet, Girshick2014rcnn, girshick15fastrcnn, RenNIPS15fasterRCNN}, including our own work, which uses the Fast R-CNN detector as its staring point~\cite{girshick15fastrcnn}. We next discuss in more detail specific innovations in this paradigm and how they relate to our approach.

\textbf{Context}: Context is known to play an important role in visual recognition~\cite{torralba2003contextual}. Numerous ideas for exploiting context in CNNs have been proposed. Sermanet et al.~\cite{sermanetCVPR2013} used two contextual regions centered on each object for pedestrian detection. In~\cite{Szegedy15}, in addition to region specific features, features from the entire image are used to improve region classification. He et al.~\cite{He2014sppNet} implement context in a more implicit way by aggregating CNN features prior to classification using different sized pooling regions. More recently, \cite{gidaris2015object} proposed to use ten contextual regions around each object with different crops. Our approach is most related to~\cite{gidaris2015object}, however, we use just four contextual regions organized in a foveal structure and importantly our classifier is trained jointly end-to-end.

\textbf{Skip connections}: Sermanet et al.~\cite{sermanetCVPR2013} proposed to use a `multi-stage' classifier that used features at many convolutional layers for pedestrian detection, showing improved results. Such `skip' architectures have recently become popular for semantic segmentation~\cite{longCVPR15fcn, BharathCVPR2015}. Concurrently with our work, Bell et al.~\cite{bell15ion} proposed to revisit skip connections for general object detection. Our own implementation of skip connections closely resembles~\cite{bell15ion}.

\textbf{Object Proposals}: When originally introduced, object proposals were based on low-level grouping cues, edges, and superpixels~\cite{AlexePAMI12, Uijlings13, ZitnickD14, CVPR2015MCG, Hosang2015proposals}. More recently, large gains in proposal quality have been achieved through use of CNNs~\cite{Szegedy15, RenNIPS15fasterRCNN, pinheiro2015learning, pinheiro2016refining}. In this work we use DeepMask segmentation proposals~\cite{pinheiro2015learning}. Specifically, we used an early version of the improved variant of DeepMask described in~\cite{pinheiro2016refining} that includes top-down refinement but is based on the VGG-A architecture~\cite{Simonyan15}, not the later ResNet architecture presented in~\cite{he2015deep}. Overall, we obtain substantial improvements in detection accuracy on COCO by using DeepMask in place of the Selective Search~\cite{Uijlings13} proposals used in the original work on Fast R-CNN~\cite{girshick15fastrcnn}.

\textbf{Classifier}: The CNN used for classification forms an integral part of the detection pipeline and is key in determining final detector accuracy. The field has witnessed rapid progress in CNNs in recent years. The introduction of AlexNet~\cite{AlexNet} reinvigorated the use of deep learning for visual recognition. The much deeper VGG~\cite{Simonyan15} and GoogleNet~\cite{GoogLeNet} models further pushed accuracy. In our work we use variants of the VGG network~\cite{Simonyan15}, specifically VGG-A for DeepMask and VGG-D for our MultiPath network. In concurrent work, He at al.~\cite{he2015deep} introduced the even deeper Residual Networks (ResNet) that have greatly improved the state of the art and have also proven effective for object detection. We expect that integration of ResNet into our system could further boost accuracy.

\section{Methods}

A high-level overview of our detection model is shown in \fig{model}. Our system is based on the Fast R-CNN framework~\cite{girshick15fastrcnn}. As in Fast R-CNN, the VGG-D network~\cite{Simonyan15} (pretrained on ImageNet~\cite{imagenet_cvpr09}) is applied to each input image and RoI-pooling is used to extract features for each object proposal. Using these features, the final classifier outputs a score for each class (plus the background) and predicts a more precise object localization via bounding box regression. We refer readers to~\cite{girshick15fastrcnn} for details.

We propose the following modifications to this basic setup. First, instead of a single classifier head, our model has four heads that observe different-sized context regions around the bounding box in a `foveal' structure. Second, each of these heads combines features from the conv3, conv4, and conv5 layers. Finally, the outputs of the four classifiers are concatenated and used to compute a score based on our proposed integral loss. Similar to Fast R-CNN, the network also performs bounding box regression using these same features.

As information can flow through several parallel pathways of our network we name it a MultiPath CNN. We describe details of each modification next.

\subsection{Foveal Structure}

Fast R-CNN performs RoI-pooling on the object proposal bounding box without explicitly utilizing surrounding information. However, as discussed, context is known to play an important role in object recognition~\cite{torralba2003contextual}. We also observed that given only cropped object proposals, identification of small objects is difficult even for humans without context.

To integrate context into our model, we looked at the promising results from the `multiregion' model~\cite{gidaris2015object} for inspiration. The multiregion model achieves improved localization results by focusing on 10 separate crops of an object with varying context. We hypothesized that this mainly improves localization from observing the object at multiple scales with increasing context, rather than by focusing on different parts of the object.

Therefore, to incorporate context, we add four region crops to our model with `foveal' fields of view of 1$\times$, 1.5$\times$, 2$\times$ and 4$\times$ of the original proposal box all centered on the object proposal. In each case we use RoI-pooling to generate features maps of the same spatial dimensions given each differently-sized foveal region. The downstream processing shares an identical structure for each region (but with separate parameters), and the output features from the four foveal classifiers are concatenated into a single long vector. This feature vector is used for both classification and bounding box regression. See \fig{model} for details.

Our foveal model can be interpreted as a simplified version of the multiregion model that only uses four regions instead of the ten in~\cite{gidaris2015object}. With the reduced number of heads, we can train the network end-to-end rather than each head separately as in \cite{gidaris2015object}.

\subsection{Skip Connections}

Fast R-CNN performs RoI-pooling after the VGG-D conv5 layer. At this layer, features have been downsampled by a factor of 16. However, 40\% of COCO objects have area less than $32\times32$ pixels and 20\% less than $16\times16$ pixels, so these objects will have been downsampled to $2\times2$ or $1\times1$ at this stage, respectively. RoI-pooling will upsample them to $7\times7$, but most spatial information will have been lost due to the $16\times$ downsampling of the features.

Effective localization of small objects requires higher-resolution features from earlier layers~\cite{sermanetCVPR2013, longCVPR15fcn, BharathCVPR2015, bell15ion, pinheiro2016refining}. Therefore, we concatenate the RoI-pooled normalized features from conv3, conv4, and conv5 layers in the same manner as described in \cite{bell15ion} and provide this as input to each foveal classifier, as illustrated in \fig{model}. A $1\times1$ convolution is used to reduce the dimension of the concatenated features to the classifier input dimension. The largest foveal features will not need as fine-grained features, so as an optimization, we sparsify these connections slightly. Specifically, we only connect conv3 to the $1\times$ classifier head and conv4 to the $1\times$, $1.5\times$, and $2\times$ heads. Overall, these skip connections give the classifier access to information from features at multiple resolutions.

\subsection{Integral Loss}

In PASCAL~\cite{Everingham10} and ImageNet~\cite{imagenet_cvpr09}, the scoring metric only considers whether the detection bounding box has intersection over union (IoU) overlap greater than 50 with the ground truth. On the other hand, the COCO evaluation metric~\cite{mscoco2015} averages AP across IoU thresholds between 50 and 95, awarding a higher AP for higher-overlap bounding boxes\footnote{Going forward, we use the notation introduced by the COCO dataset~\cite{mscoco2015}. Specifically, we use AP to denote AP averaged across IoU values from 50 to 95, and \AP{u} to denote AP at IoU threshold $u$ (e.g., the PASCAL metric is denoted by \AP{50}). Note also that we use the convention that IoU ranges from 0 to 100.}. This incentivizes better object localization. Optimizing \AP{50} has resulted in models that perform basic object localization well but often fail to return tight bounding boxes around objects.

For training, Fast R-CNN uses an IoU threshold of 50. We observed that changing this foreground/background threshold $u$ during training improves \AP{u} during testing, but can decrease AP at other IoU thresholds. To target the integral AP, we propose a loss function that encourages a classifier to perform well at multiple IoU thresholds.

The original loss $L$ used in Fast R-CNN~\cite{girshick15fastrcnn} is given by:
\begin{equation}
 L(p, k^*, t, t^*) = L_\textrm{cls}(p, k^*)
 + \lambda [k^* \ge 1] L_\textrm{loc}(t, t^*),
\end{equation}%
for predicted class probabilities $p$, true class $k^*$, predicted bounding box $t$, and true bounding box $t^*$. The first term $L_\textrm{cls}(p, k) = -\log p_{k^*}$ is the classification log loss for true class $k^*$. The second term, $L_\textrm{loc}(t,t^*)$, encourages the class-specific bounding box prediction to be as accurate as possible. The combined loss is computed for every object proposal. If the proposal overlaps a ground truth box with IoU greater than 50, the true class $k^*$ is given by the class of the ground truth box, otherwise $k^*=0$ and the second term of the loss is ignored.

Observe that in the original R-CNN loss, the classification loss $L_\textrm{cls}$ does not prefer object proposals with high IoU: all proposals with IoU greater than 50 are treated equally. Ideally, proposals with higher overlap to the ground truth should be scored more highly. We thus propose to modify $L_\textrm{cls}$ to explicitly measure integral loss over all IoU thresholds $u$:
\begin{equation}
 \int_{50}^{100}L_\textrm{cls}(p, k^*_u)du,
\end{equation}%
where $k^*_u$ is the true class at overlap threshold $u$. We approximate this integral as a sum with $du=5$ and modify our network to output multiple corresponding predictions $p_u$. Specifically, our modified loss can be written as:
\begin{equation}
 L(p, k^*, t, t^*) = \frac{1}{n}\sum_u \Big[ L_\textrm{cls}(p_u, k^*_u)
 + \lambda [k^*_u \ge 1] L_\textrm{loc}(t, t^*) \Big].
\end{equation}%
We use $n=6$ thresholds $u \in\{50,55,\ldots,75\}$. Note that in this formulation each object proposal actually has $n$ ground truth labels $k^*_u$, one label per threshold $u$. In our model, each term $p_u$ is predicted by a separate head, see \fig{model}. Specifically, for each $u$, we train a separate linear classifier (using shared features) to predict the true class $k^*_u$ of a proposal (where the ground truth label is defined using threshold $u$). At inference time, the output softmax probabilities $p_u$ of each of the $n$ classifiers are averaged to compute the final class probabilities $p$. The modified loss function and updated network encourages object proposals with higher overlap to the ground truth to be scored more highly.

During training, each head has progressively fewer total positive training samples as there are fewer proposals overlapping the ground truth as $u$ is increased. To keep the ratio of sampled positive and negative examples constant for each head, each minibatch is constructed to train a single head in turn. We restrict the heads to the range $u\le75$, otherwise the proposals contain too few total positive samples for training. Finally, note that for bounding box regression, our network is unchanged and predicts only a single bounding box output $t$.

\section{Experiments}\label{sec:results}

In this section we perform a detailed experimental analysis of our MultiPath network. For all following experiments, Fast R-CNN~\cite{girshick15fastrcnn} serves as our baseline detector (with VGG-D~\cite{Simonyan15} features pre-trained on ImageNet~\cite{imagenet_cvpr09}). We use DeepMask object proposals~\cite{pinheiro2015learning, pinheiro2016refining} and focus exclusively on the recent COCO dataset~\cite{mscoco2015} which presents novel challenges for detection.

We begin by describing the training and testing setup in \S\ref{sec:results:setup}. Next, in \S\ref{sec:results:main} we study the impact of each of our three core network modifications, including skip connections, foveal regions, and the integral loss. We analyze the gain from DeepMask proposals in \S\ref{sec:results:deepmask} and compare with the state of the art in \S\ref{sec:competition}. Finally, in the appendix we analyze a number of key parameters and also additional modifications that by and large did \emph{not} improve accuracy. 

Our system is written using the Torch-7 framework. All source code for reproducing the methods in this paper will be released.

\subsection{Training and Testing Setup}\label{sec:results:setup}

For all experiments in this section we report both the overall AP (averaged over multiple IoU thresholds) and \AP{50}. All our models are trained on the 80K images in COCO 2014 train set and tested on the first 5K images from the val set. We find that testing on these 5K images correlates well with the full 40K val set and 20K test-dev set, making these 5K images a good proxy for model validation without the need to test over the full val or test-dev sets.

Training is performed for 200K iterations with 4 images per batch and 64 object proposals per image. We use an initial learning rate of $10^{-3}$ and reduce it to $10^{-4}$ after 160K iterations. Training the full model takes \app3 days on 4 NVIDIA Titan X GPUs. Unless noted, in testing we use a non maximal suppression threshold of 30, 1000 proposals per image, an image scale of 800 pixels, and no weight decay (we analyze all settings in the appendix).

Both data and model parallelism are used in training~\cite{owt}. First, 4 images are propagated through the VGG-D network trunk, in parallel with 1 image per GPU. The features are then concatenated into one minibatch and subsequently used by each of the 4 foveal regions. Each foveal region resides in a separate GPU. Note that the prevalence of 4 GPU machines helped motivate our choice of using 4 foveal regions due to ease of parallelization.

Our network requires 150ms to compute the features and 350ms to evaluate the foveal regions, for a total of about 500ms per COCO image. We time with a scale of 800px and 400 proposals (see appendix and \fig{proposals}). Fast R-CNN with these settings is about $2\times$ faster.

\subsection{MultiPath Network Analysis}\label{sec:results:main}

\begin{table}[t]\centering\scriptsize
\renewcommand\arraystretch{1.1}\renewcommand{\tabcolsep}{2mm}
\begin{tabular}[t]{c c c | c c}
 integral loss & foveal & skip & \AP{50} & AP \\
 \shline
        &      &      & 43.4 & 25.2\\
   \chk &      &      & 42.2 & 25.6\\
        & \chk &      & 45.2 & 25.8\\
   \chk & \chk &      & 44.4 & 26.9\\
        & \chk & \chk & \textbf{46.4} & 27.0 \\
   \chk & \chk & \chk & 44.8 & \textbf{27.9} \\
\end{tabular}\hspace{10mm}
\begin{tabular}[t]{c c c | c c}
  integral loss & context & \#regions & \AP{50} & AP \\
  \shline
        & none        & 1  & 43.4 & 25.2 \\
        & multiregion & 10 & 44.0 & 25.5 \\
        & foveal      & 4  & \textbf{45.2} & \textbf{25.8} \\
  \hline
   \chk & none        & 1  & 42.2 & 25.6 \\
   \chk & multiregion & 10 & 43.1 & 26.3 \\
   \chk & foveal      & 4  & \textbf{44.4} & \textbf{26.9}
\end{tabular}
\Caption{\textbf{Left:} Model improvements of our MultiPath network. Results are shown for various combinations of modifications enabled. Each contributes roughly equally to final accuracy, and in total AP increases 2.7 points to 27.9. \textbf{Right:} Our 4-region foveal setup versus the 10 regions used in multiregion~\cite{gidaris2015object}. Surprisingly, our approach outperforms~\cite{gidaris2015object} despite using fewer regions. See text for details.}
\label{table:main}
\end{table}

Our implementation of Fast R-CNN~\cite{girshick15fastrcnn} with DeepMask object proposals~\cite{pinheiro2015learning} achieves an overall AP of 25.2 and an \AP{50} of 43.4. This is already a large improvement over the original Fast R-CNN results that used Selective Search proposals~\cite{Uijlings13}, we will return to this shortly.

A breakdown of how each of our three core network modifications affects AP and \AP{50} over our strong baseline is shown in \tab{main}, left. Results are shown for each combination of modifications enabled including skip connections, foveal regions, and the integral loss (except skip connections were implemented only for foveal regions). Altogether, AP increases 2.7 points to 27.9, with each modification contributing \app1 point to final performance. \AP{50} improves 1.4 points to 44.8; however, not surprisingly, the best \AP{50} of 46.4 is achieved without the integral loss. We carefully analyze the foveal structure and integral loss next.

\textbf{Foveal structure}: A breakdown of the gains from using foveal regions is shown in \tab{main}, right, both with and without the integral loss but without skip connections. Gains from foveal regions are amplified when using the integral loss, resulting in an AP improvement of 1.3 points. We also compare our foveal approach to the multiregion network~\cite{gidaris2015object} which used 10 regions (for a fair comparison, we re-implement it in our setup). Surprisingly, it performs slightly \emph{worse} than our foveal setup despite having more regions. This may be due to the higher number of parameters or it's possible that this requires more iterations to converge.

\begin{figure}[t]\centering
 \includegraphics[width=0.49\textwidth]{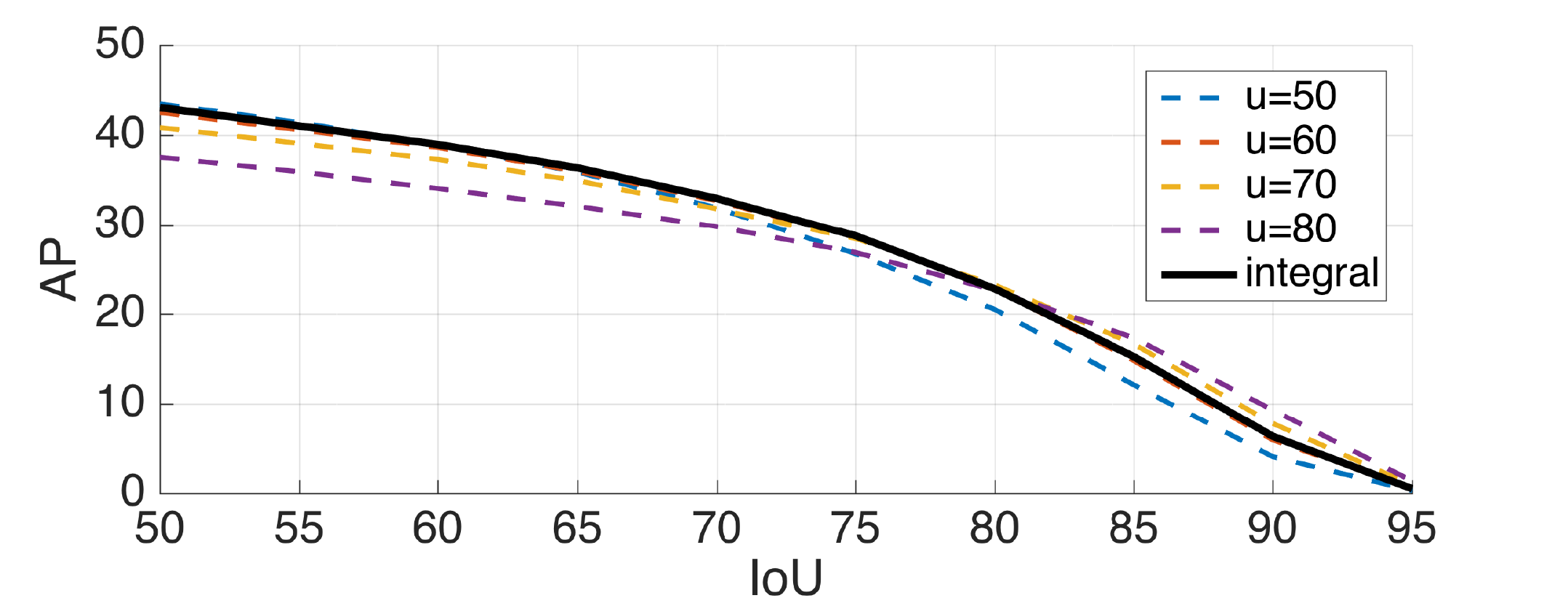}
 \includegraphics[width=0.49\textwidth]{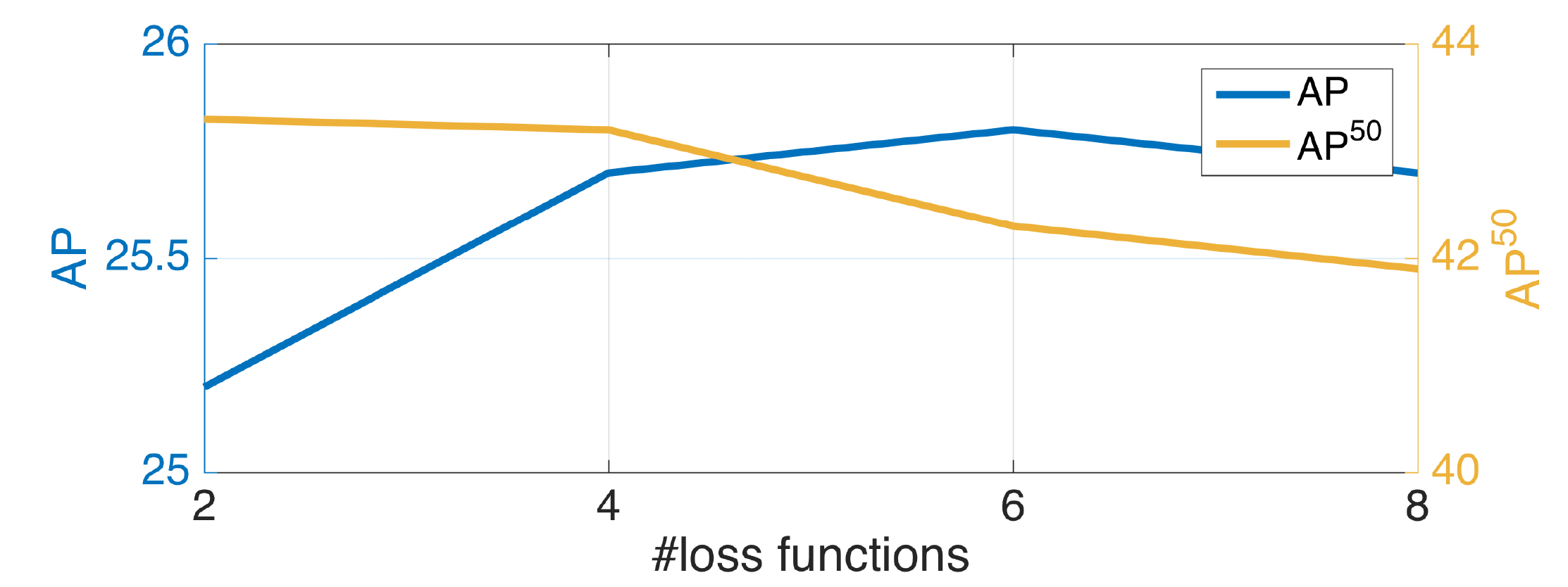}
\Caption{\textbf{Left}: Each standard model performs best at the threshold used for training while using the integral loss yields good results at all settings. \textbf{Right}: Integral loss achieves best AP with 6 heads.}
\label{fig:integral}
\end{figure}

\textbf{Integral Loss}: \fig{integral}, left, shows AP at various IoU thresholds for models trained with different IoU cutoffs $u$ as well as our integral loss. Each standard model tends to perform best at the IoU for which it was trained. Integral loss improves overall AP by \app1 over the $u=50$ model, and does so while maintaining a slightly higher \AP{50} than simply increasing $u$ (e.g. our \AP{50} is 0.6 points higher than the $u=60$ model). \fig{integral}, right, shows AP and \AP{50} for varying number of heads. Using 6 heads $(u\le75)$ achieves the highest AP. For the experiments in \fig{integral} we trained for 280K iterations as we found the integral loss requires somewhat longer to converge (we used 200K iterations for all other ablations studies).

\subsection{DeepMask Proposals}\label{sec:results:deepmask}

\begin{figure}[t]\centering
 \includegraphics[width=.49\textwidth]{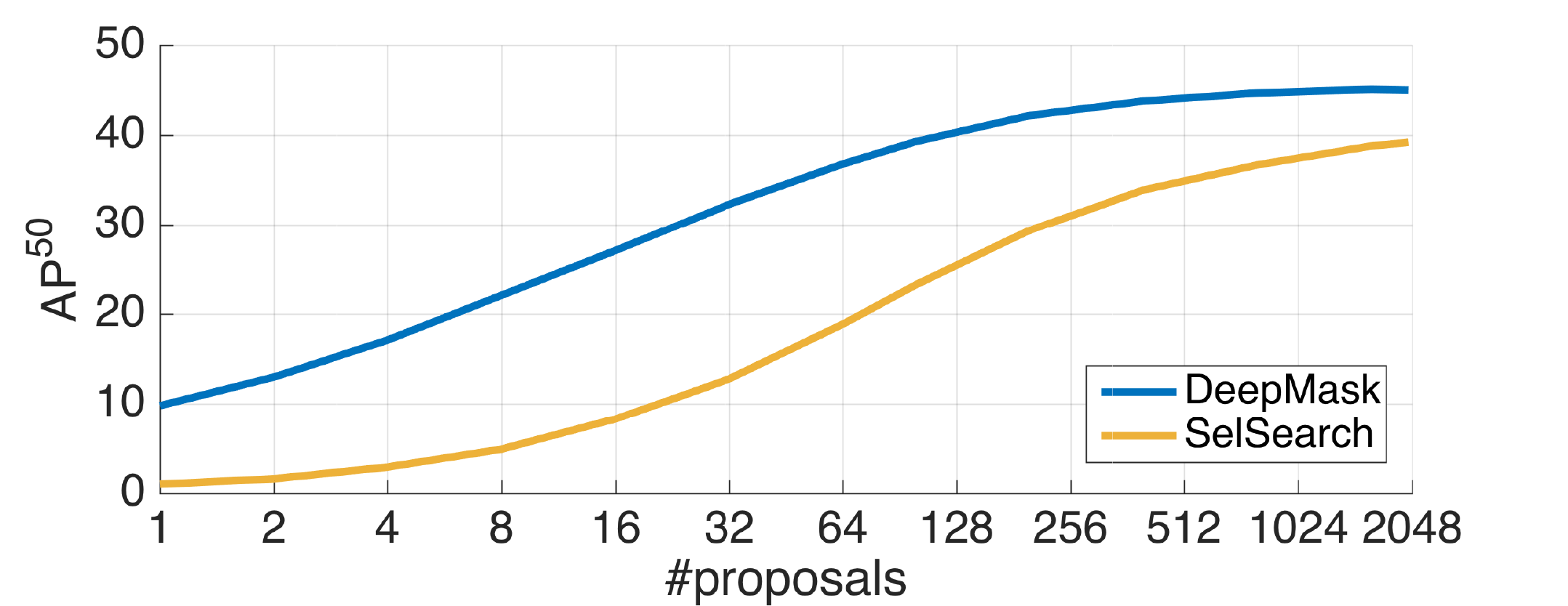}
 \includegraphics[width=.49\textwidth]{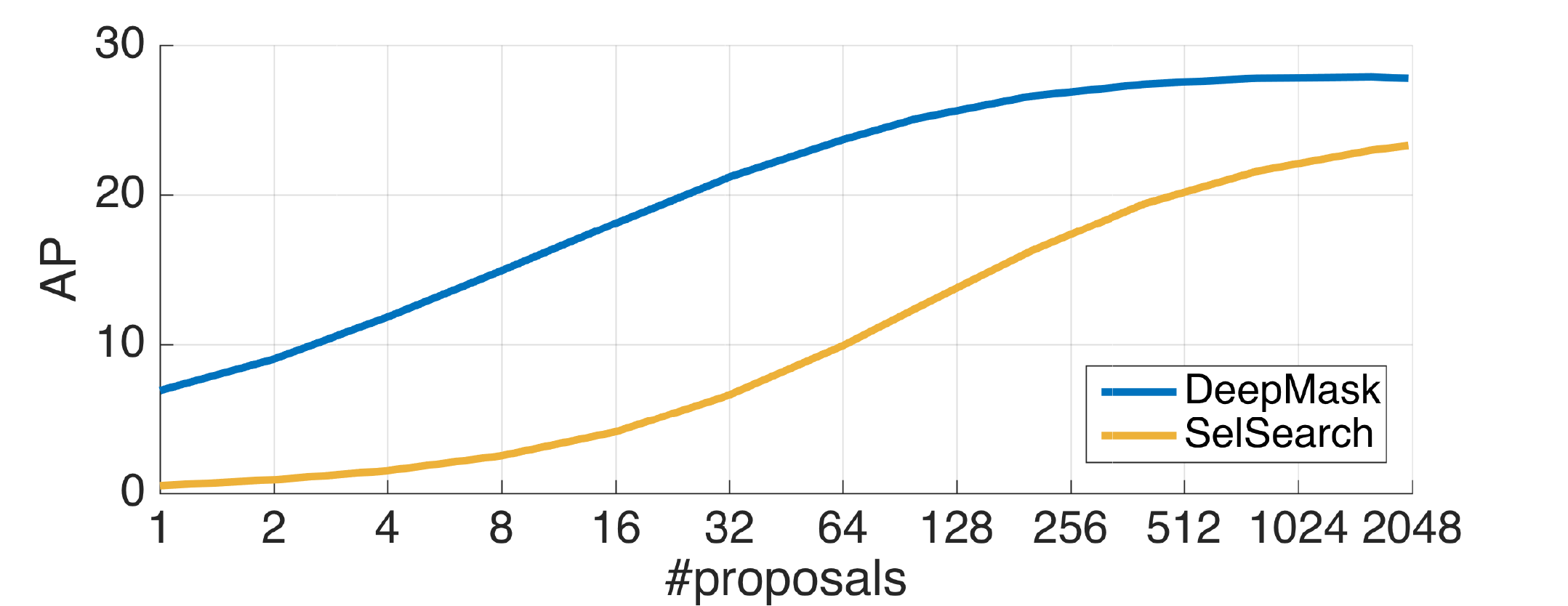}
\Caption{\AP{50} and AP versus number and type of proposals. Accuracy saturates using 400 DeepMask proposals per image and using \app50 DeepMask proposals matches 2000 Selective Search proposals.}
\label{fig:proposals}
\end{figure}

Object proposals play a central role in determining detector accuracy. The original implementation of Fast R-CNN with Selective Search proposals~\cite{Uijlings13} has an AP of 19.3. Our MultiPath network improves this to 22.8 AP using these same proposals. Switching to DeepMask proposals~\cite{pinheiro2015learning, pinheiro2016refining} increases accuracy by a further very substantial 5.1 points to 27.9 AP.

\fig{proposals} shows \AP{50} and AP for varying number and type of proposals. Not only is accuracy substantially higher using DeepMask, fewer proposals are necessary to achieve top performance. Our results saturate with around 400 DeepMask proposals per image and using just 50 DeepMask proposals matches accuracy with 2000 Selective Search proposals.

Interestingly, our setup substantially reduces the benefits provided by bounding box regression. With the original Fast R-CNN and Selective Search proposals, box regression increases AP by 3.5 points, but with our MultiPath model and DeepMask proposals, box regression only increases AP by 1.1 points. See \tab{improvements}, left, for details.

\begin{table}[t]\centering\scriptsize
\renewcommand\arraystretch{1.1}\renewcommand{\tabcolsep}{1.8mm}
\begin{tabular}[t]{ l | c c c | c c c}
 &  \multicolumn{3}{c|}{\AP{50}} & \multicolumn{3}{c}{AP} \\
 & base & +bb & $\Delta$ & base & +bb & $\Delta$ \\
 \shline
 SS + Fast R-CNN & 38.2 & 39.8 & +1.6 & 18.1 & 21.6 & +3.5\\
 SS + MultiPath  & 38.0 & 38.5 & +0.5 & 20.9 & 22.8 & +1.9\\
 \hline
 DM + Fast R-CNN & 42.5 & 43.4 & +0.9 & 23.5 & 25.2 & +1.7\\
 DM + MultiPath  & 44.5 & 44.8 & +0.3 & 26.8 & 27.9 & +1.1\\
\end{tabular}\hspace{6mm}
\begin{tabular}[t]{ l | c c | c c}
  & \AP{50} & $\Delta$ & AP & $\Delta$ \\
 \shline
  baseline            & 44.8 &      & 27.9 &  \\
  + trainval   & 47.5 & +2.7 & 30.2 & +2.3 \\
  + hflip      & 48.3 & +0.8 & 30.8 & +0.6 \\
  + FMP        & 49.6 & +1.3 & 31.5 & +0.7 \\
  + ensembling & 51.9 & +2.3 & 33.2 & +1.7 \\
\end{tabular}
\Caption{\textbf{Left:} Bounding box regression is key when using Selective Search (SS) proposals and the Fast R-CNN classifier (our implementation). However, with DeepMask (DM) proposals and our MultiPath network, box regression increases AP by only 1.1 points (and \AP{50} by 0.3) as our pipeline already outputs well-localized detections. \textbf{Right:} Final enhancements to our model. Use of additional training data, horizontal flip at inference, fractional max pooling (FMP), and ensembling gave a major cumulative boost. These are common approaches for maximizing accuracy, see appendix for details.}
\label{table:improvements}
\end{table}

\section{COCO 2015 Results}\label{sec:competition}

\begin{figure}[t]\centering
 \includegraphics[width=\textwidth]{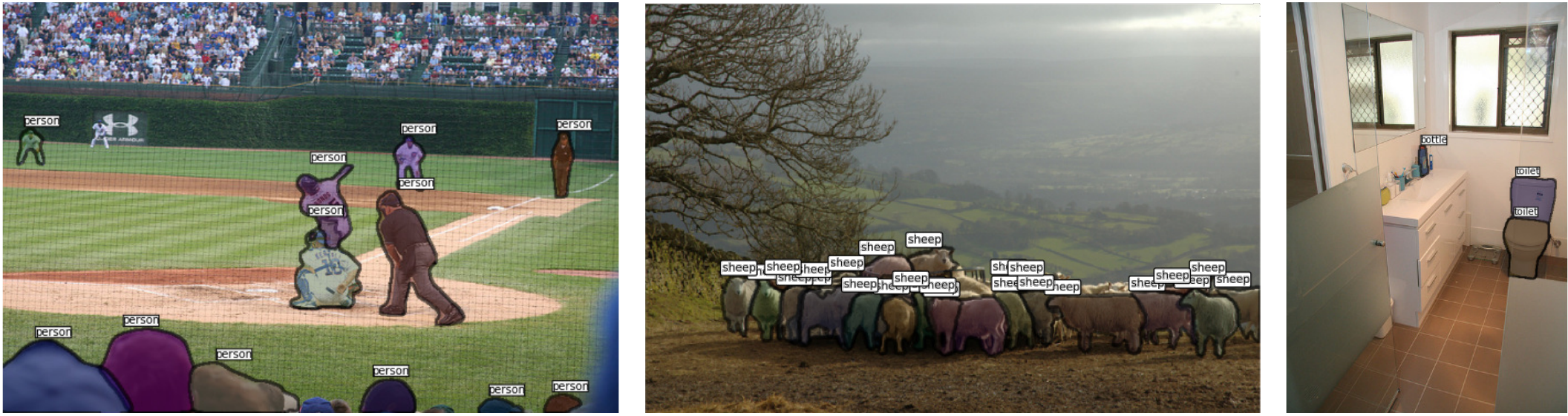}
\Caption{Selected detection results on COCO. Only high-scoring detections are shown. While there are missed objects and false positives, many of the detections and segmentations are quite good.}
\label{fig:results3}
\end{figure}

\begin{table}[t]\centering\scriptsize
\renewcommand\arraystretch{1.1}\renewcommand{\tabcolsep}{.9mm}
\begin{tabular}{l|cccccccccccc}
  & AP & AP$^{50}$ & AP$^{75}$ & AP\tss{S} & AP\tss{M} & AP\tss{L} &
  AR$^1$ & AR$^{10}$ & AR$^{100}$ & AR\tss{S} & AR\tss{M} & AR\tss{L} \\
 \shline
   ResNet \cite{he2015deep} & 27.9 & 51.2 & 27.6 & 8.6 & 30.2 & 45.3 & 25.4 & 37.1 & 38.0 & 16.6 & 43.3 & 57.8\\
   \textbf{MultiPath} & 25.0 & 45.4 & 24.5 & 7.2 & 28.8 & 39.0 & 23.8 & 36.6 & 38.5 & 17.0 & 46.7 & 53.5\\
 \hline
   ResNet \cite{he2015deep}
    & 37.1 & 58.8 & 39.8 & 17.3 & 41.5 & 52.5 & 31.9 & 47.5 & 48.9 & 26.7 & 55.2 & 67.9\\
   \textbf{MultiPath}
    & 33.2 & 51.9 & 36.3 & 13.6 & 37.2 & 47.8 & 29.9 & 46.0 & 48.3 & 23.4 & 56.0 & 66.4\\
   ION \cite{bell15ion}
    & 30.7 & 52.9 & 31.7 & 11.8 & 32.8 & 44.8 & 27.7 & 42.8 & 45.4 & 23.0 & 50.1 & 63.0\\
   Fast R-CNN* \cite{girshick15fastrcnn}
    & 19.3 & 39.3 & 19.9 &  3.5 & 18.8 & 34.6 & 21.4 & 29.5 & 29.8 &  7.7 & 32.2 & 50.2\\
   Faster R-CNN* \cite{RenNIPS15fasterRCNN}
    & 21.9 & 42.7 &  --- &  --- &  --- &  --- &  --- &  --- &  --- &  --- &  --- &  ---\\
\end{tabular}
\Caption{\textbf{Top}: COCO test-standard segmentation results. \textbf{Bottom}: COCO test-standard bounding box results (top methods only). Leaderboard snapshot from 01/01/2016. *Note: Fast R-CNN and Faster R-CNN results are on test-dev as reported in~\cite{RenNIPS15fasterRCNN}, but results between splits tend to be quite similar.}
\label{table:leaderboard}
\end{table}

To maximize accuracy prior to submitting to the COCO leaderboard, we added validation data to training, employed horizontal flip and fractional max pooling~\cite{fmp} at inference, and ensembled 6 models. Together, these four enhancements boosted AP from 27.9 to 33.2 on the held-out validation images, see \tab{improvements}, right. More details are given in the appendix. Finally, to obtain segmentation results, we simply fed the bounding box regression outputs back to the DeepMask segmentation system. Note that as discussed in \S\ref{sec:results:deepmask}, box regression only improved accuracy slightly. In principle, we could have used the original DeepMask segmentation proposals without box regression; however, we did not test this variant.

We submitted our results the COCO 2015 Object Detection Challenge. Our system placed second in both the bounding box and segmentation tracks. \tab{leaderboard} compares our results to the top COCO 2015 challenge systems and additional baselines. Only the deeper ResNet classifier~\cite{he2015deep} outperformed our approach (and potentially ResNet could be integrated as the feature extractor in our MultiPath network, leading to further gains). Compared to the baseline Fast R-CNN, our system showed the largest gains on small objects and localization, improving AP on small objects by $4\times$ and \AP{75} by 82\%.

\begin{figure}[t]\centering
 \includegraphics[width=\textwidth]{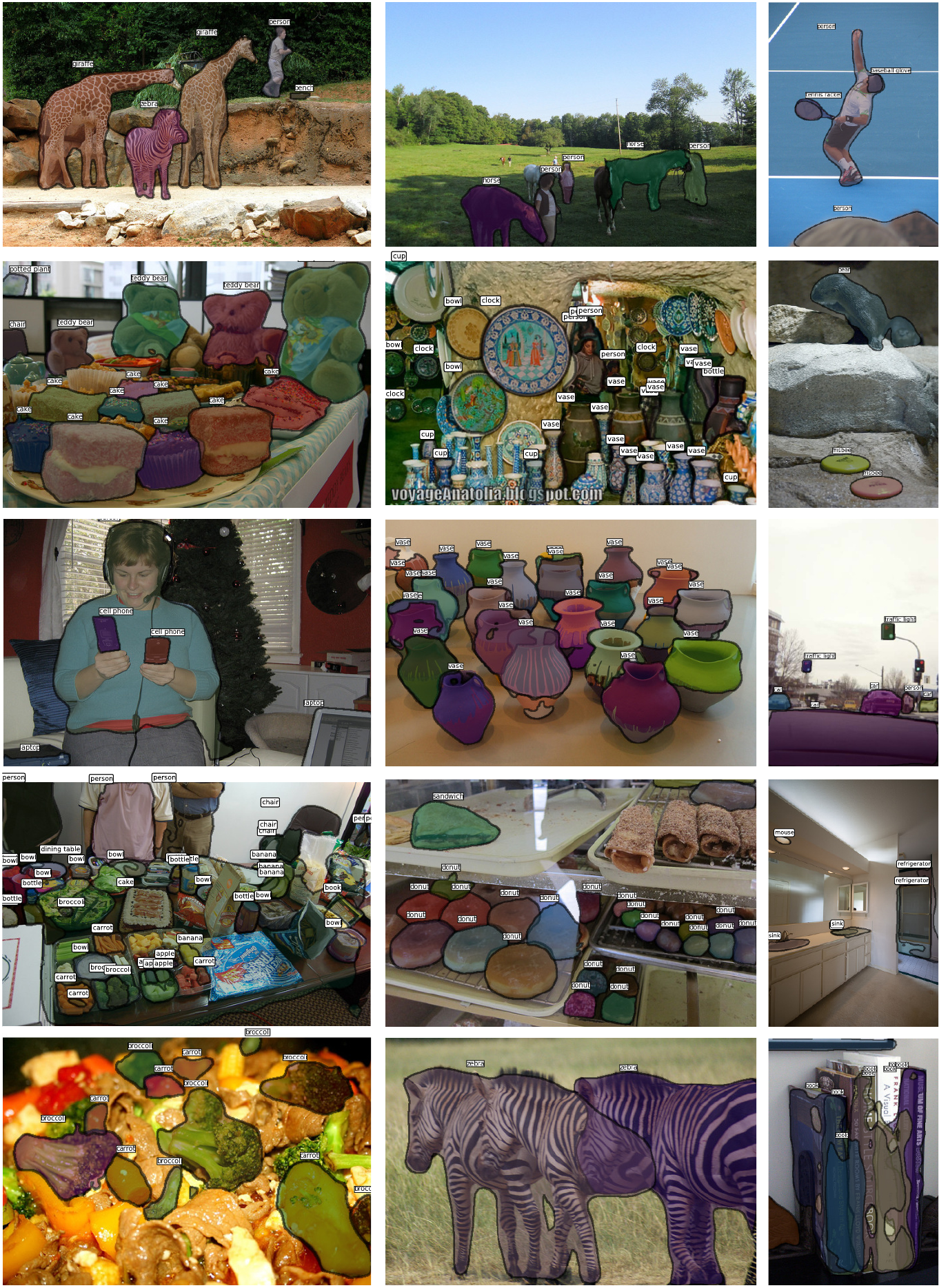}
\Caption{Selected detection results on COCO. Only high-scoring detections are shown. While there are missed objects and false positives, many of the detections and segmentations are quite good.}
\label{fig:results}
\end{figure}

\fig{results3} and \fig{results} show selected detection results from our system. \fig{analysis} shows a breakdown of errors of our system. Most of the overall error comes from false positives and negatives, with little inter-class classification error. Despite our improvements on small objects, small object detection remains quite challenging.

\section{Conclusion}

In this paper, we proposed three modifications to Fast R-CNN: (1) skip connections to give the network access to multi-scale features, (2) foveal regions to provide context, and (3) the integral loss to improve localization. We coupled our resulting MultiPath classifier with DeepMask proposals and achieved a 66\% improvement over the baseline Fast R-CNN with Selective Search. All source code for our approach will be released. Our hope is that our model can serve as a baseline system on COCO and prove useful to the detection community.

{\vspace{2mm}\hspace{-1.5em}\small\textbf{Acknowledgements}: We would like to thank Ross Girshick, Rob Fergus, Bharath Hariharan, Spyros Gidaris, Nikos Komodakis, and Francisco Massa for helpful discussion and suggestions.}

\clearpage

\newcommand{\incp}[2]{\begin{subfigure}[b]{0.315\textwidth}\centering
 \includegraphics[width=\textwidth,trim={0 0 0 .77cm},clip]{figures/analyze/#1}
 \vspace{-5mm}\Caption{#2}\label{fig:analyze:#1}\end{subfigure}}
\begin{figure}[t]\centering
 \incp{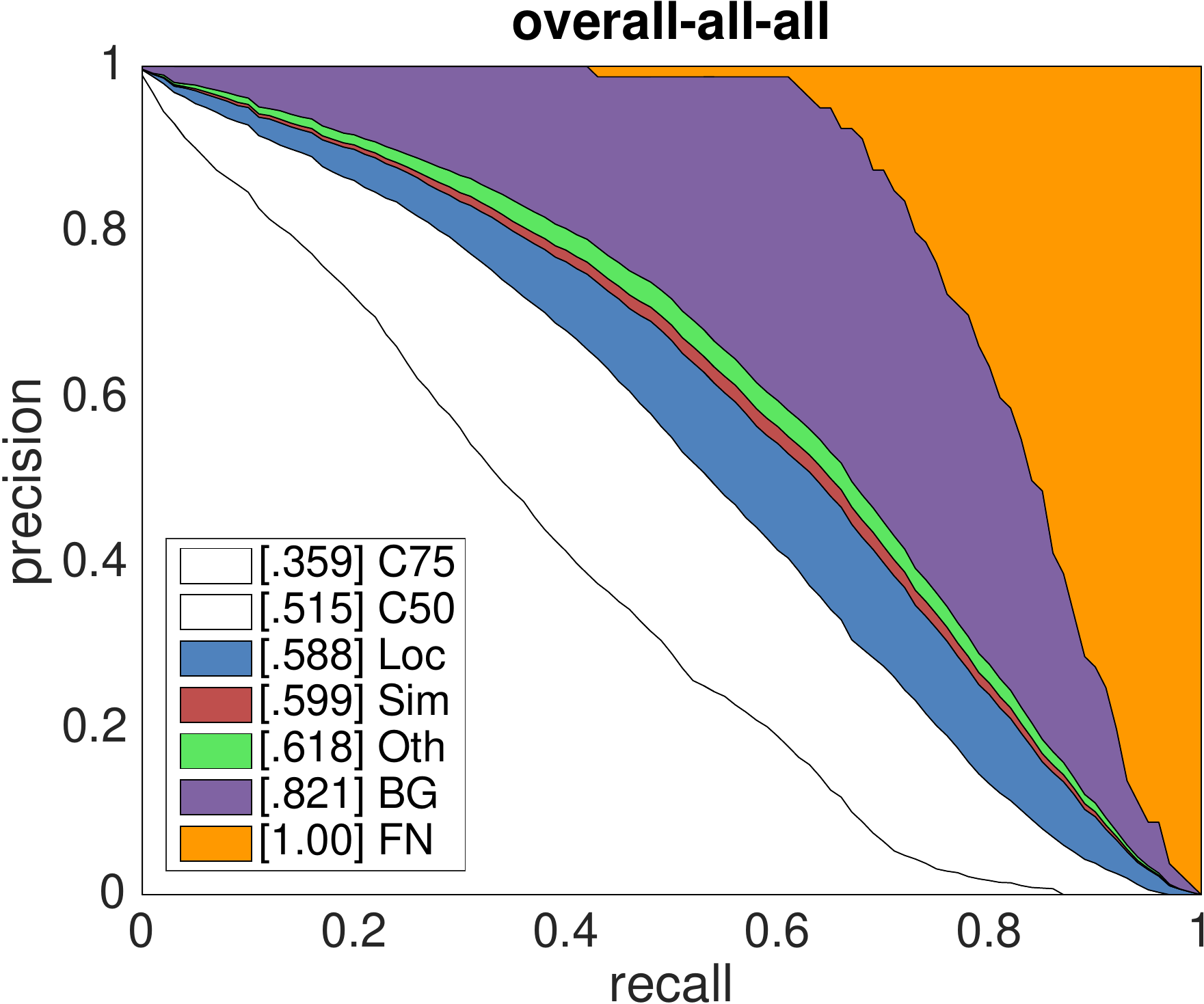}{overall}\hspace{2mm}
 \incp{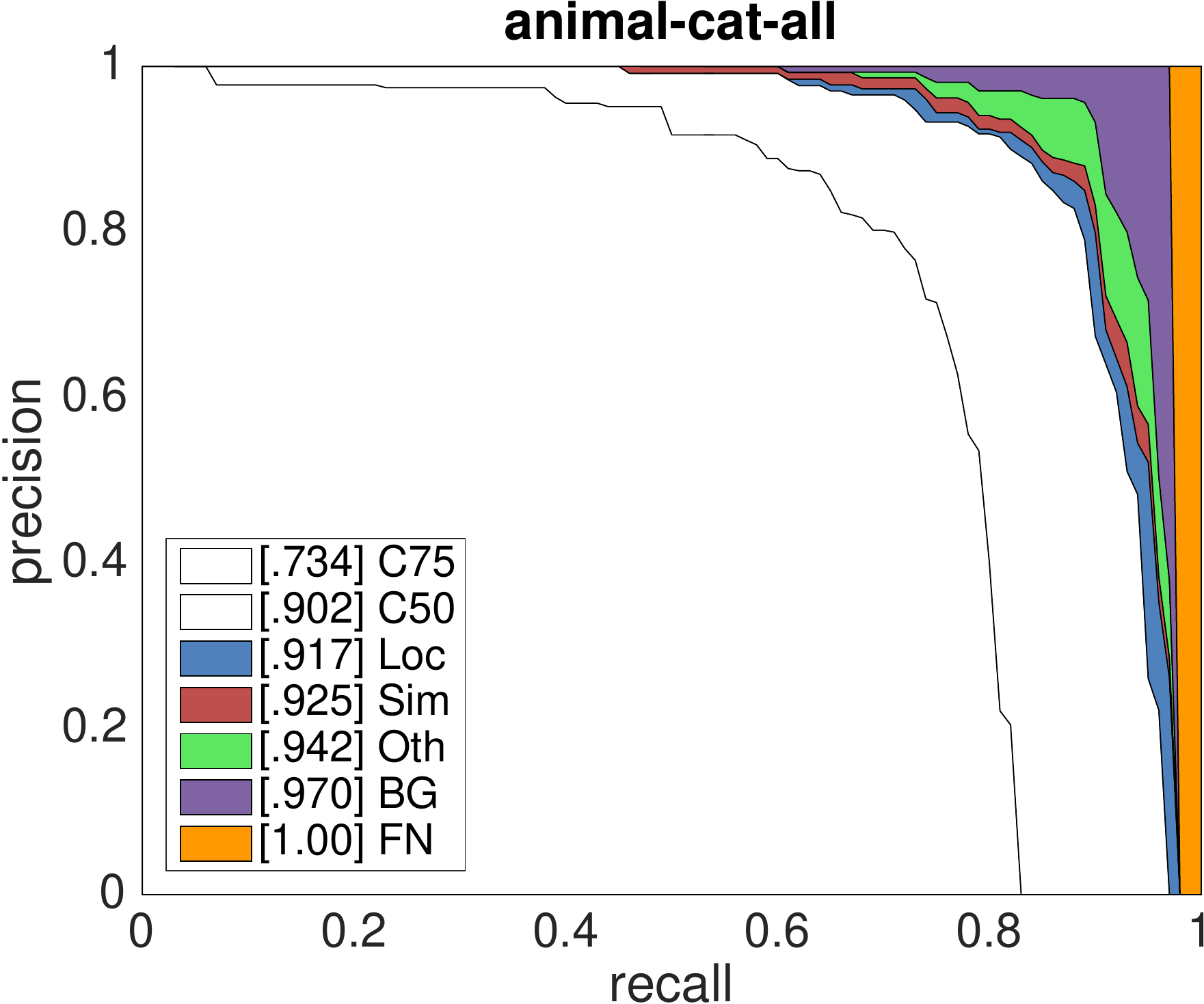}{cats}\hspace{2mm}
 \incp{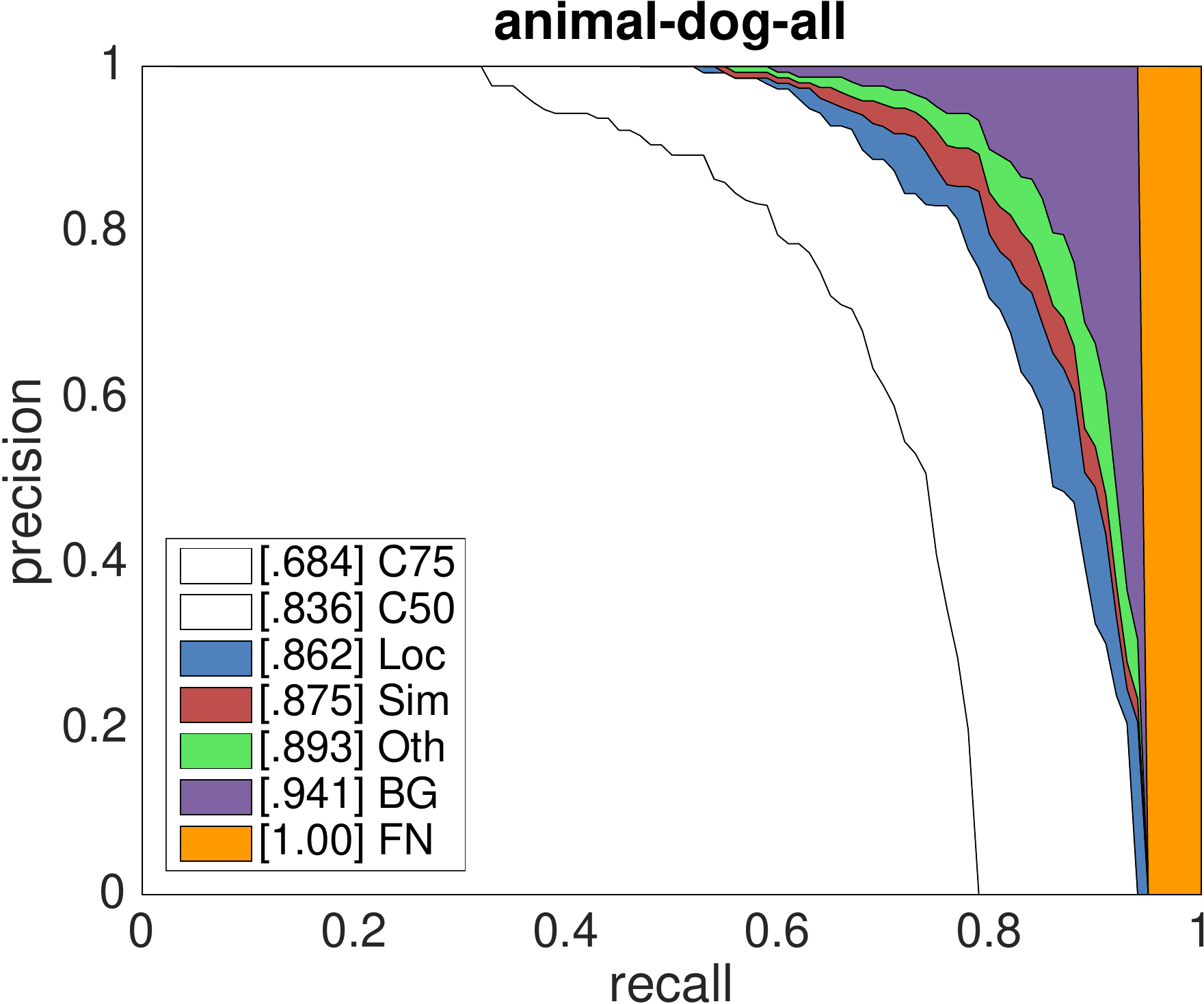}{dogs}\vspace{2mm}\\
 \incp{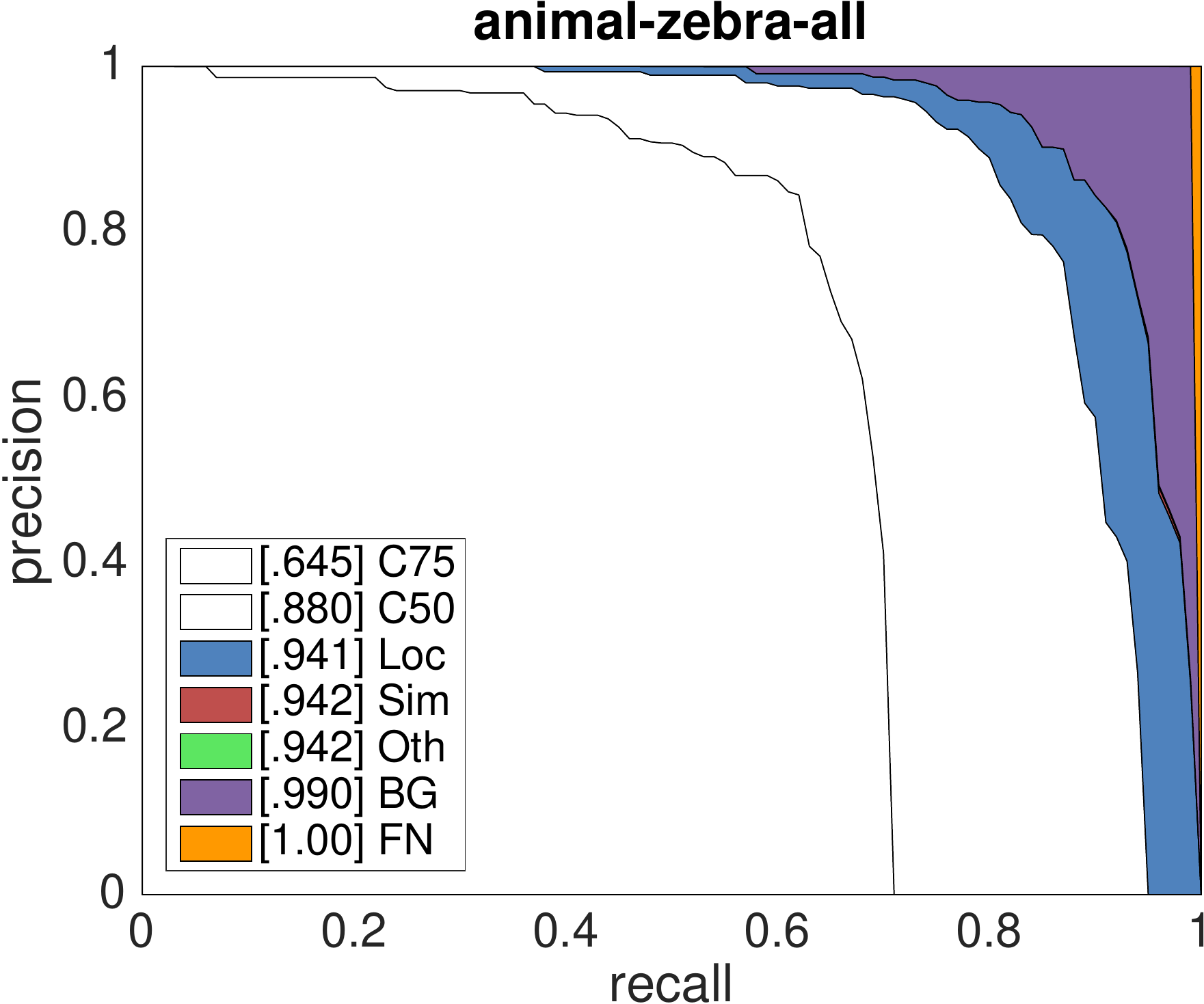}{zebras}\hspace{2mm}
 \incp{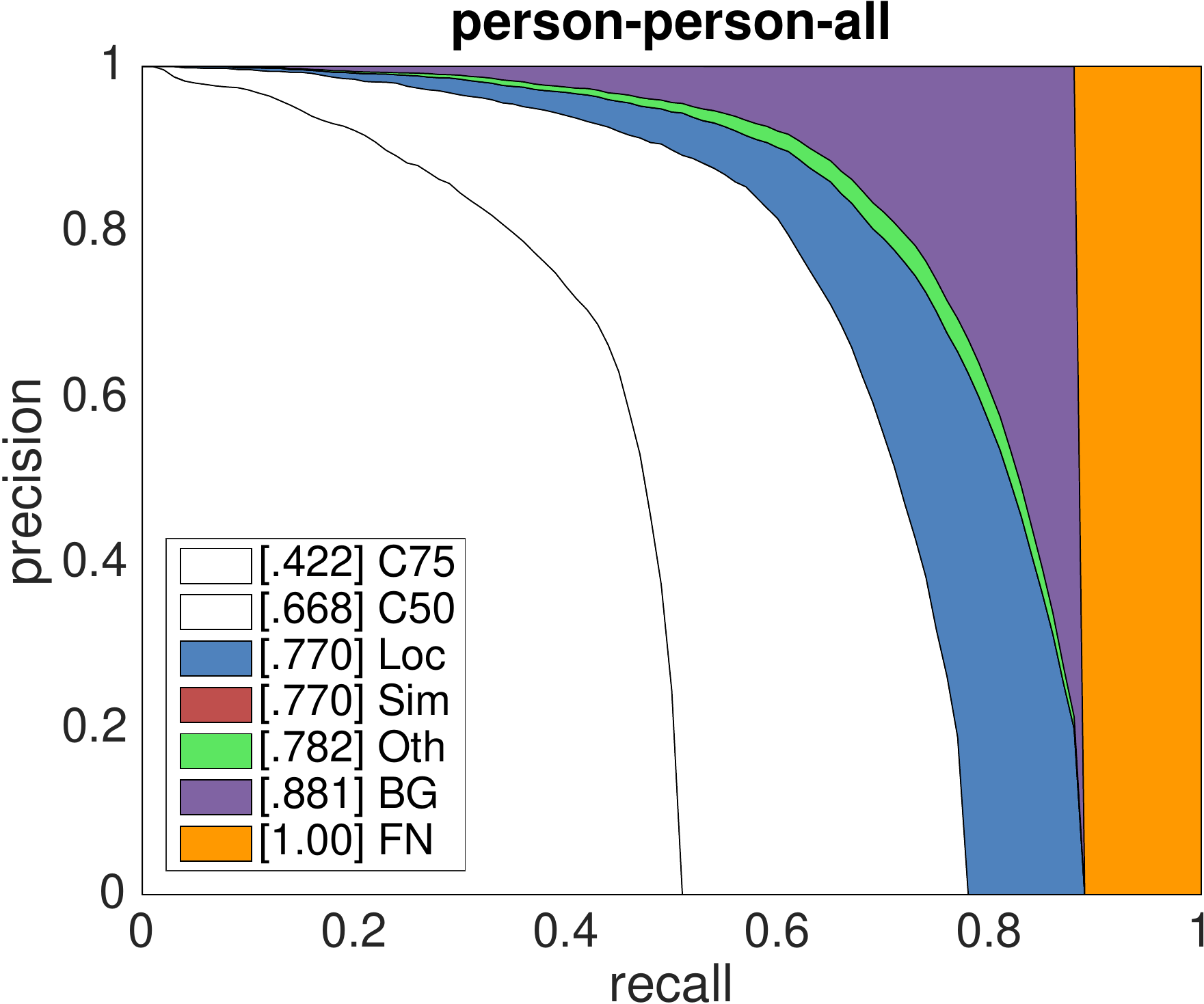}{people}\hspace{2mm}
 \incp{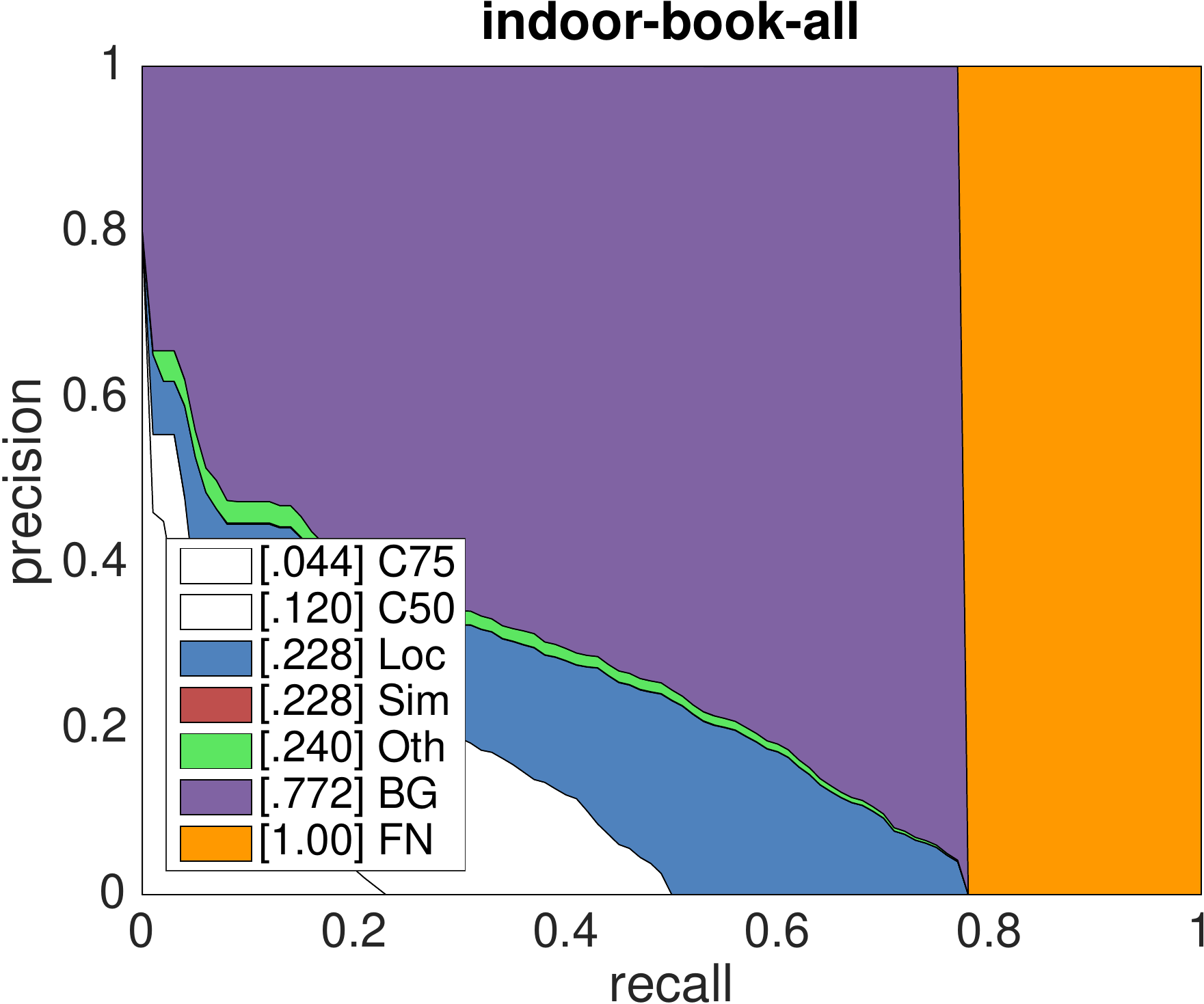}{books}\vspace{2mm}\\
 \incp{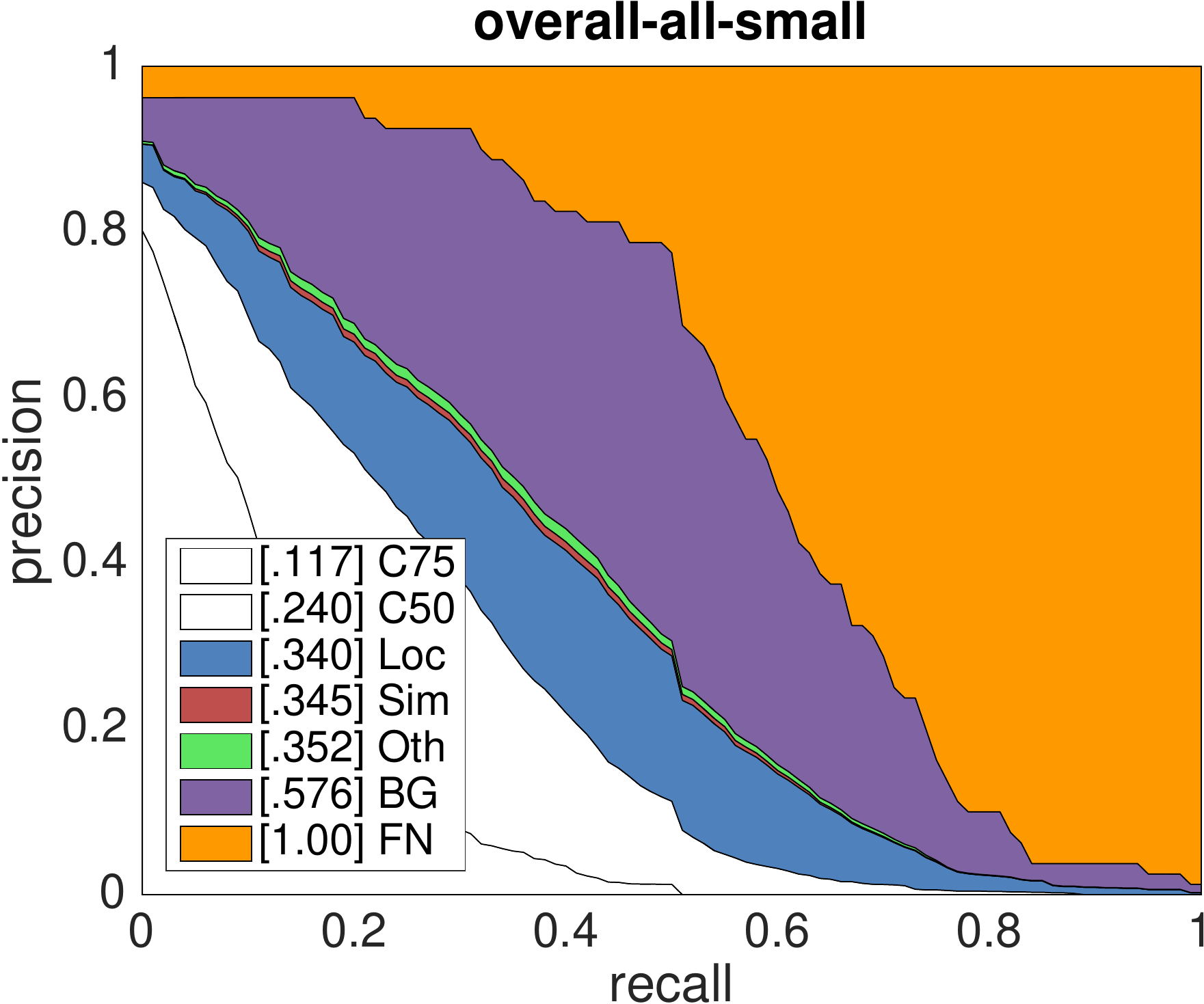}{small}\hspace{2mm}
 \incp{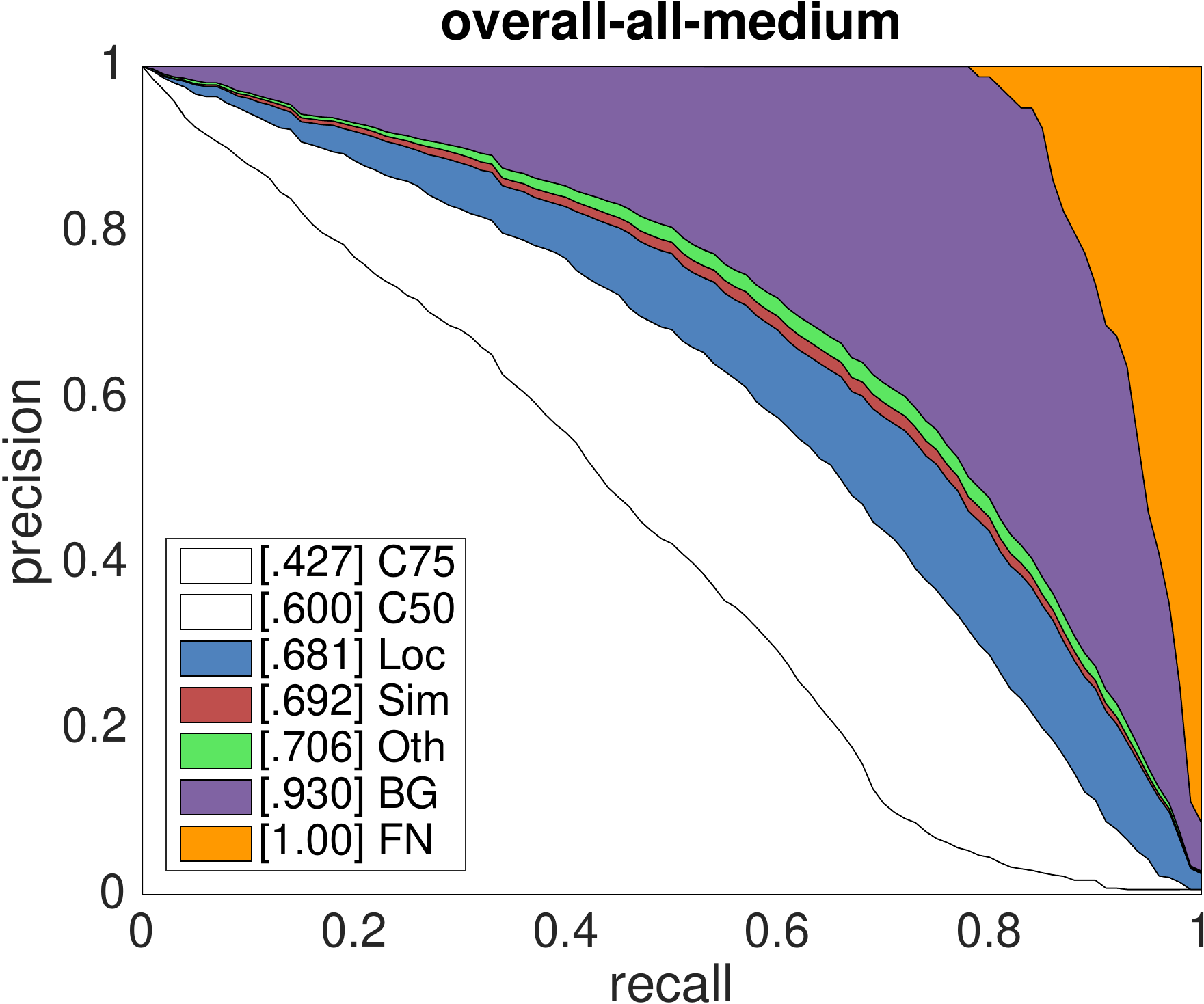}{medium}\hspace{2mm}
 \incp{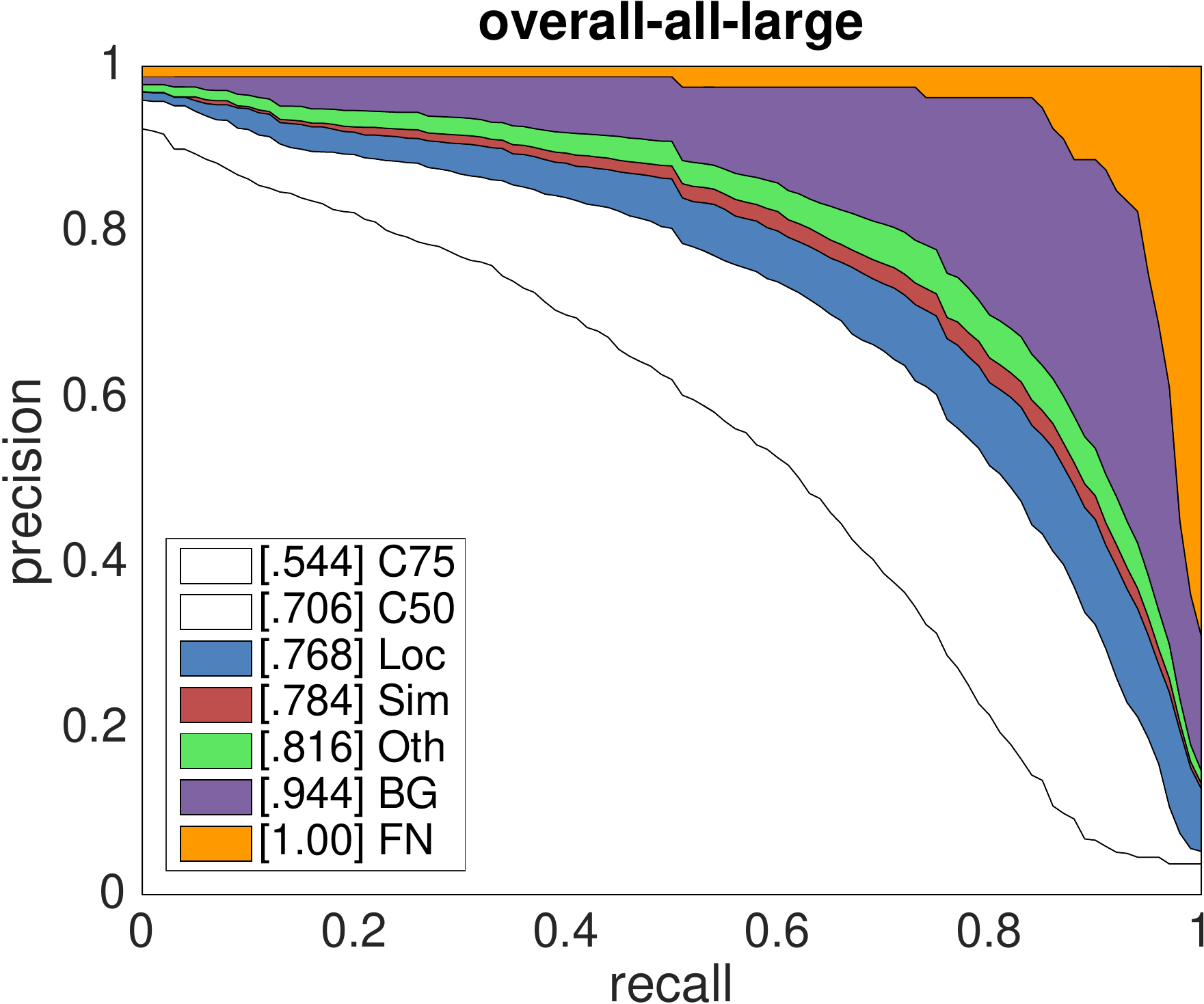}{large}\hspace{2mm}
\Caption{Detailed analysis of detector performance on unseen COCO validation images at select settings (plots in style of \cite{hoiem2012diagnosing} generated by COCO API code). {\bf(a)} Removing localization errors would lead to an \AP{10} of 58.8 on COCO (`Loc'). Removing similar and other class confusion (`Sim' and `Oth') would only lead to slight improvements in accuracy. The remaining errors are all based on background confusions (`BG') and false negatives (`FN'). {\bf(b,c)} Our detector performs similarly on cats and dogs, achieving high overall accuracy with some class and background confusions but few missed detections. {\bf(d)} Zebras are quite distinct, however, localization of overlapping zebras can be difficult due to their striped patterns. {\bf(e)} People are the dominant category on COCO and have average difficulty. {\bf(f)} Books are an incredibly difficult category due to their small size and highly inconsistent annotation in COCO. {\bf(g,h,i)} Accuracy broken down by scale; not unexpectedly, small objects ($area<32^2$) are quite difficult, while accuracy on large objects ($area>96^2$) is much higher. While there is a practical limit to the performance on small objects which are often ambiguous or poorly-labeled, there is still substantial opportunity for improvement. We expect better proposals, more accurate filtering of false positives, and stronger reasoning about context can all improve small object detection.}
\label{fig:analysis}
\end{figure}
\clearpage

\section*{Appendix: Additional Analysis}\label{sec:appendix}

\begin{figure}[t]\centering
 \includegraphics[width=.49\textwidth]{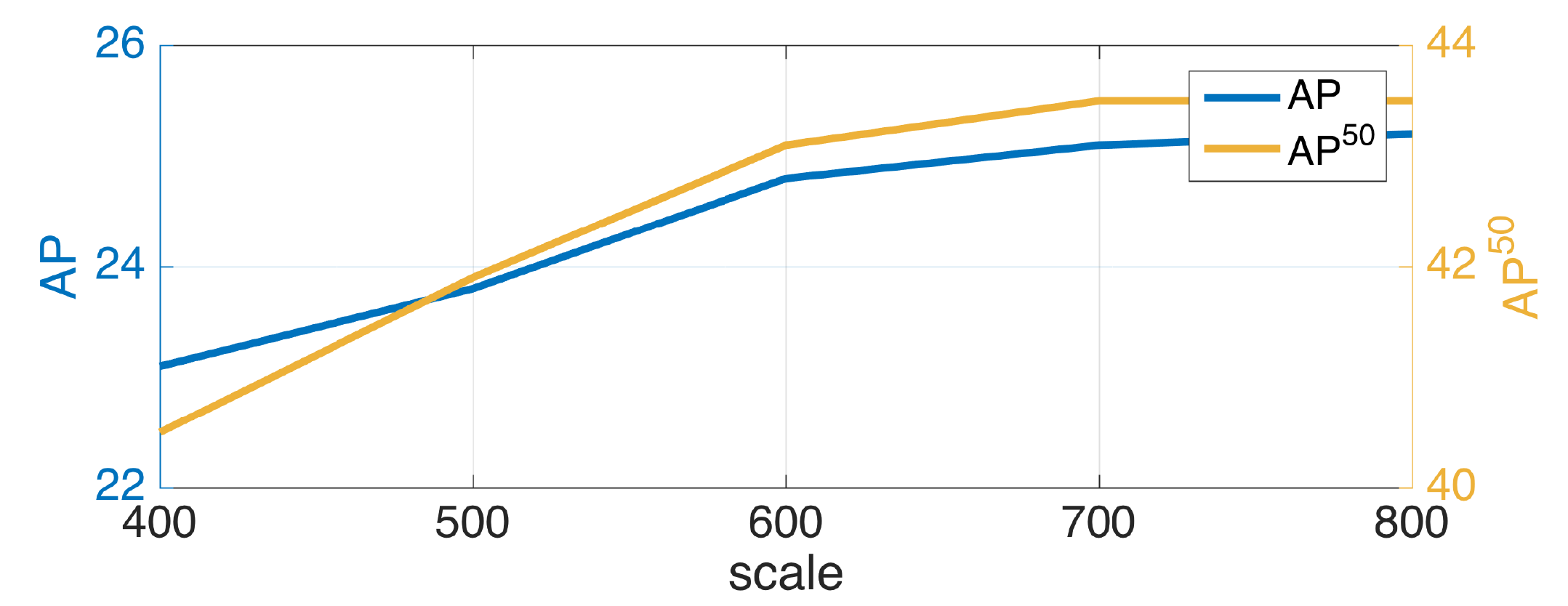}
 \includegraphics[width=.49\textwidth]{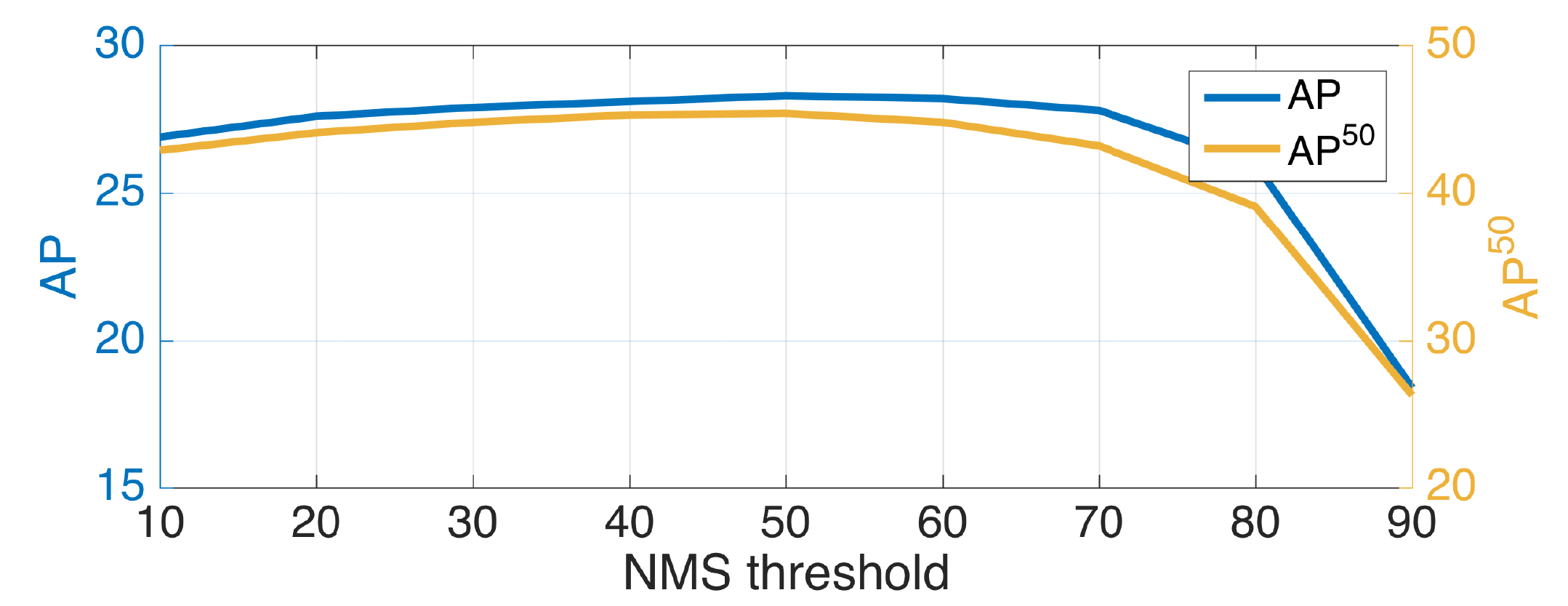}
\Caption{Effect of scale (left) and NMS threshold (right) on detection performance.\vspace{-3mm}}
\label{fig:scale+nms}
\end{figure}

In the appendix we describe our additional enhancements reported in \tab{improvements} and analyze a number of key parameters. We also report additional modifications that did \emph{not} improve accuracy; we hope that sharing our \emph{negative results} will prove beneficial to the community.

\textbf{train+val}: Adding validation data to training (minus the 5K held-out images from the validation set we use for testing) improved accuracy by 2.3 points AP, see \tab{improvements}. We trained for 280K iterations in this case. We note that the DeepMask proposals were only trained using the train set, so retraining these on train+val could further improve results.

\textbf{hflip}: Fast R-CNN is not invariant to horizontal image flip (hflip) even though it is trained with hflip data augmentation. Thus, we average the softmax scores from the original and flipped images and also average the box regression outputs (directly, not in log space). AP improves by 0.6 points, see \tab{improvements}.

\textbf{FMP}: Inspired by Fractional Max Pooling~\cite{fmp}, we perform multiple RoI-pooling operations with perturbed pooling parameters and average the softmax outputs (note that the network trunk is computed only once). Specifically, we perform two ROI-poolings: the first follows \cite{He2014sppNet} and uses the floor and ceil operations for determining the RoI region, the second uses the round operation. As shown in \tab{improvements}, FMP improves AP 0.7 points.

\textbf{Ensembling}: Finally, we trained an ensemble of 6 similar models. Each model was initialized with the same ImageNet pre-trained model, only the order of COCO training images changed. This ensemble boosted AP 1.7 points to 33.2, see \tab{improvements}.

\textbf{Scale}: \fig{scale+nms}, left, shows accuracy as a function of image scale (minimum image dimension in pixels with maximum fixed to 1000px). Increasing scale improves accuracy up to \app800px, but at increasing computation time. We set the scale to 800px which improves AP by 0.5 points over the 600px scale used by \cite{girshick15fastrcnn} for PASCAL.

\textbf{NMS threshold}: \fig{scale+nms}, right, shows accuracy as a function of the NMS threshold. Fast R-CNN~\cite{girshick15fastrcnn} used a threshold of 30. For our model, an NMS threshold of 50 performs best, improving AP by 0.4 points, possibly due to the higher object density in COCO.

\textbf{Dropout \& Weight Decay}: Dropout helped regularize training and we keep the same dropout value of 0.5 that was used for training VGG-D. On the other hand, setting weight decay to 0 for fine-tuning improved results by 1.1 \AP{50} and 0.5 AP. Note that \cite{bell15ion} used weight decay but not dropout, so perhaps it is sufficient to have just one form of regularization.

\textbf{Iterative Localization}: Bounding box voting with iterative localization as proposed in \cite{gidaris2015object} did \emph{not} substantially improve the AP of our model, again probably due to the higher quality of DeepMask proposals and the improved localization ability of our MultiPath network.

\textbf{ImageNet Data Augmentation}: As there are some under-represented classes in COCO with few annotations, we tried to augment the training set with ImageNet 2012 detection training data. Surprisingly, this only improved performance on the most underrepresented class: hair dryer; for all other classes, accuracy remained unchanged or suffered.

\bibliography{bibliography}

\end{document}